\documentclass[letterpaper, 10 pt, conference]{ieeeconf}  
\IEEEoverridecommandlockouts                              

\usepackage{graphics} 
\usepackage{epsfig} 
\usepackage{times} 
\usepackage{amsmath} 
\usepackage{amssymb}  
\usepackage{float}
\usepackage[caption = false]{subfig}
\usepackage{afterpage}
\usepackage[table]{xcolor}
\usepackage{booktabs}
\usepackage{caption}
\usepackage{amsfonts}
\usepackage{algorithm,algpseudocode}

\usepackage{balance}
\usepackage{dblfloatfix}
\title{\LARGE \bf
Multi-Agent Ergodic Exploration under Smoke-Based Time-Varying Sensor Visibility Constraints
}
\newcommand{\norm}[1]{\left\lVert#1\right\rVert}

\author{Elena Wittemyer$^1$, Ananya Rao$^2$, Ian Abraham$^1$, and Howie Choset$^2$
\thanks{$^1$Yale University, Intelligent Autonomy Lab}
\thanks{$^2$Carnegie Mellon University, Robotics Institute}
}

\begin{document}

\maketitle
\thispagestyle{empty}
\pagestyle{empty}

\begin{abstract}
In this work, we consider the problem of multi-agent informative path planning (IPP) for robots whose sensor visibility continuously changes as a consequence of a time-varying natural phenomenon. We leverage ergodic trajectory optimization (ETO), which generates paths such that the amount of time an agent spends in an area is proportional to the expected information in that area. We focus specifically on the problem of multi-agent drone search of a wildfire, where we use the time-varying environmental process of smoke diffusion to construct a sensor visibility model. This sensor visibility model is used to repeatedly calculate an expected information distribution (EID) to be used in the ETO algorithm. Our experiments show that our exploration method achieves improved information gathering over both baseline search methods and naive ergodic search formulations.
\end{abstract}
\section{INTRODUCTION}
The use of drones in exploration of extreme natural environments has become popular due to their relatively low cost and ability to reach areas too unsafe for humans \cite{daud2022applications}. By ``extreme natural environments," we mean natural settings with demanding physical phenomena that complicate the process of exploration, such as extreme temperature variations or destructive weather events. We use multi-drone systems for this problem of exploration of extreme natural environments as they speed up the exploration process and provide critical redundancies in case of damage or loss.


However, the dynamic nature of such environments often complicates the process of path planning for multi-drone systems. Time-varying factors, like weather events or moving targets, can change both the distribution of information in the exploration space as well as the capabilities of the multi-agent team. As such, we must account for time-varying environmental factors in the path planning algorithm in order for drones to successfully explore extreme environments. In this work, we focus on path planning with regard to the environmental factor of changing visibility constraints. Moreover, we demonstrate the effectiveness of our visibility-aware planning method on environments with dynamic information distributions.

In informative path planning (IPP) problems, we encode environmental changes in environmental process models, which describe the evolution of a process in the exploration area over time \cite{ouyang2014multi}. Much prior work on IPP focuses on how information or obstacles in an exploration space will change according to an environmental process model~\cite{karur2021survey}. However, given that robots are not isolated from the changes in the environment they are exploring, it is equally critical to consider how the effectiveness of a robot's sensor will change with some environmental process. Issues including decreased power availability, visibility occlusions, exposure to hazardous chemicals, or extreme temperatures can all dramatically reduce the performance of a robot's sensor \cite{osha}. 

\begin{figure}[t]
    \centering
    \includegraphics[width=\linewidth]{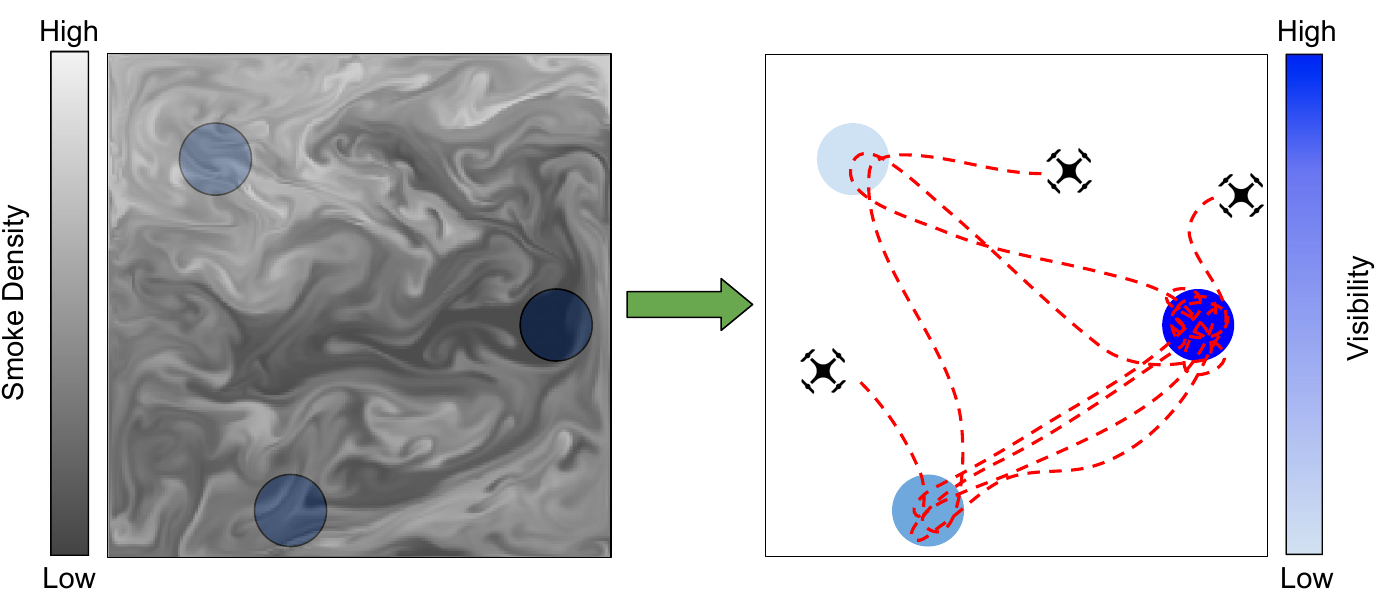}
    \caption{\textbf{Multi-Agent Exploration with Varying Visibility Constraints.} Three information peaks have varying visibility according to the concentration of smoke over the peaks. Given that less information can be captured when observing peaks with low visibility, the multi-drone team prioritizes exploring peaks with high visibility.}
    \label{fig:fig_1}
\end{figure}

We consider the motivating problem of multi-drone exploration in a wildfire environment. The frequency and severity of wildfires has increased in recent years \cite{westerling2016increasing}, and so the use of autonomous robots for wildfire monitoring or search and rescue is particularly critical~\cite{twidwell2016smokey}. For both of these problems, the aim of the multi-drone team is to gain confidence in the predicted state (position, velocity, etc) of a set of targets (humans, buildings, fire sources, etc). To accomplish this, agents take sensor measurements that reduce uncertainty in the target belief state. However, the smoke emitted by the wildfire can occlude the drones' visual sensors, lessening their ability to capture new images and thus reduce uncertainty. 

In this work, we model the drones' changing sensor effectiveness by calculating expected visibility according to the density of smoke in an area. We then use multi-agent ergodic trajectory optimization (ETO) to plan paths for drones over a map of uncertainty in a set of target states; this uncertainty map can change over time both due to measurement by drones and due to movement of the underlying targets. We leverage the smoke model alongside this known uncertainty map to calculate an expected information distribution (EID) used by the ETO algorithm. The EID encodes the predicted uncertainty reduction achieved by measuring at each location in the environment.

In ETO, we generate trajectories from an underlying information distribution so that the amount of time a robot spends in a region is proportional to the distribution of information in that region. As such, ETO balances exploitation of high-information regions with coverage of the exploration space. This is a valuable characteristic for wildfire exploration, as it both enables drones to take advantage of a-priori information for efficient information gain while still exploring low-uncertainty areas which may contain unforeseen, valuable information. An example of the trajectories generated by ETO over a visibility-based EID is shown in Fig.~\ref{fig:fig_1}.

The primary contributions of this paper are as follows:
\begin{enumerate}
    \item Improved information gathering through integration of visibility constraints into the expected information distribution used by the ETO algorithm;
    \item Increased efficiency in achieving ergodic coverage  by using Shannon entropy to calculate expected information from a model of a natural phenomenon;
    \item An extension of visibility-aware, multi-agent ETO to dynamic information maps.
\end{enumerate}

\begin{figure*}
    \centering \includegraphics[width=\linewidth]{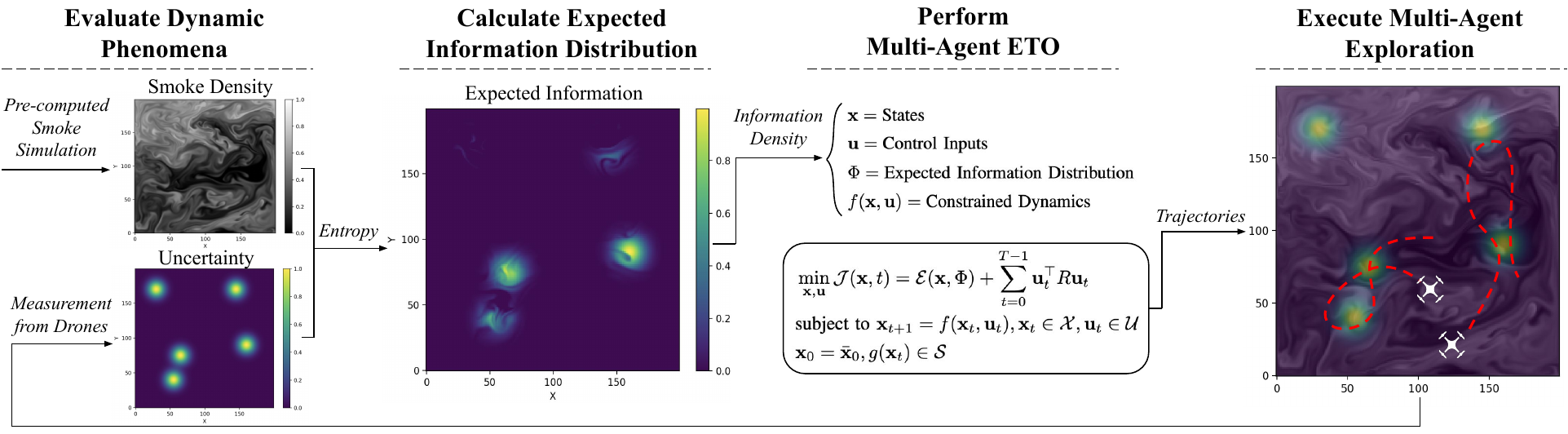}
    \caption{\textbf{Time-Varying, Visiblity-Aware Multi-Agent Ergodic Exploration.} Shown is a flow chart of our visibility-aware, time-varying ergodic exploration process. To begin, the underlying uncertainty distribution and smoke model are used to calculate the EID. Using this distribution, multi-agent ETO is performed. The agents follow these generated trajectories, and the uncertainty map is updated using the agents' measurements.}
    \label{fig:fig_2}
\end{figure*}

\section{RELATED WORK}
\noindent
\textbf{Informative Path Planning:}
Using information acquired in an environment to inform path planning is a widely investigated problem with applications to mapping, navigation, and localization \cite{bai2021information}. In much prior work, the goal of IPP is to find and measure near the highest concentration of some spatial process \cite{marchant2014sequential} \cite{blanchard2022informative} \cite{singh2009nonmyopic}. This approach is useful for environmental process monitoring and feature detection. 
For the task of exploration, in which the robot system seeks to widely observe some spatially-distributed phenomenon, a more common goal of information-theoretic planning is to minimize entropy/uncertainty using sensor observations at optimally-selected locations \cite{584097}, \cite{jiang2019maximum}, \cite{popovic2017online}, \cite{cao2013multi}. For the application of wildfire exploration, uncertainty reduction is a useful goal as the state of targets in the search environment is generally highly uncertain. 

Some prior works that use IPP for uncertainty reduction leverage Bayesian Optimization (BO) \cite{marchant2014sequential} \cite{bai2016information}. This approach iteratively samples from an objective to find a maximum without consideration of how an agent's movement constraints affect possible sampling locations. Many works use optimal rapidly-exploring random trees (RRT*) \cite{karaman2011sampling} for IPP problems \cite{schmid2020efficient} \cite{hollinger2013sampling}, which find paths from start nodes to target nodes with regard to a control-effort cost function. More advanced than RRT* is Monte Carlo Tree Search (MCTS) \cite{kodgule2019non} \cite{9196631}, which generates long-term measurement strategies by considering potential future measurements. Graph-based search methods, which discretize the exploration spaces into graphs and visit nodes within these graphs, have also widely been used for IPP problems \cite{binney2010informative} \cite{zhu2021online}. For the problem of wildfire exploration, we are interested in a planning method that does not solely prioritize information gain (like BO) or exploration space coverage (like graph-based methods), that does not neglect the movement constraints of a robot in planning sampling locations (like MCTS), and that does not require defined target node positions to plan paths (like RRT*) as the exploration space will continuously change.

For the sake of this work, we choose to use ergodic search. As previously described, in ergodic search, the time-averaged trajectories of a robotic system are planned to be proportional to an information distribution \cite{mathew2011metrics}. The benefits of ergodic search in comparison to other IPP methods are that it balances exploitation of high-entropy regions with coverage of low-entropy regions, improving exploration outcomes, and that it accounts for the robot's dynamics constraints, generating dynamically feasible trajectories.

\textbf{Path Planning for Dynamic Maps:}
Many prior works have focused on the problem of exploration of maps whose information or target distribution changes over time \cite{singh2009nonmyopic} \cite{liu2018novel} \cite{yao2015real}. 
\cite{ouyang2014multi} presents a multi-agent exploration algorithm for predicting a dynamic environmental phenomenon modeled as a Dirichlet process mixture of multiple Gaussian processes. The authors aim to plan optimal sampling locations for prediction of a spatially-varying phenomenon. Meanwhile, we seek to leverage a known spatially-varying phenomenon to better plan a series of sampling locations proportional to an uncertainty distribution. 

Ergodic search has been extended to the problem of exploring dynamic information maps \cite{miller2015ergodic} \cite{candela2021using} \cite{wittemyer2023bi}. \cite{coffin2022multi} uses multi-agent ergodic search to localize a single moving target by finding measurements that have a high probability of reducing uncertainty about target location. Our work also focuses on gaining knowledge of a target through measurements that reduce uncertainty in the state of the target, but we extend this approach to maps containing multiple static and moving targets. \cite{rao2023multi} proposes a Dynamic Multi-Objective Ergodic Search (D-MO-ES) algorithm for single agent exploration of dynamic information maps with multiple dynamic objectives. Our work extends exploration over dynamic uncertainty to multiple agents.

\textbf{Planning over Time-Varying Sensor Constraints:}
\cite{binney2013optimizing} plans optimal waypoints for measurement of a spatial phenomenon in a time-varying field, in which the utility of a measurement captured at a location is time-dependent. \cite{ghaffari2018gaussian} proposes a Gaussian Processes occupancy mapping method that uses mutual information  calculated from future map predictions based on a 2D range-finder sensor model. \cite{popovic2020informative} introduces a terrain monitoring approach with an altitude-dependent sensor model in which uncertainty increases with camera distance to a target. 
Each of these works uses a variation of greedy search, in which the planning method attempts to find the optimal sampling location for information gain. For our application, this greedy approach could be unreliable as any incorrect assumptions about the environmental process model could cause the EID to not reflect the true distribution of information in the environment, leading informative regions erroneously absent from the EID to be entirely missed by greedy search. Our method overcomes this issue by encouraging coverage alongside exploitation, increasing the odds that unexpected regions of useful information will be discovered.

In \cite{miller2015ergodic}, Miller and Murphey propose an ergodic exploration of distributed information (EEDI) algorithm that plans trajectories for robots with nonlinear sensor dynamics. The EEDI algorithm is tested on a fish-inspired robot which uses electrolocation as a sensing technique. The authors use Fisher information to calculate predicted measurement utility from sensor efficacy at a point and re-plan trajectories according to the evolving expected information distribution. The EEDI algorithm is tested on a single robot for the task of localization of stationary targets. Our work extends the problem of ergodic exploration with time-varying sensor constraints to the task of multi-agent exploration of moving targets.
\section{PRELIMINARY BACKGROUND}
\subsection{Multi-Agent Ergodic Trajectory Optimization}
To begin, for a system of $N$ robots, we define the robots' states at time $t$ as $\bold{x}_t = \{x_1, x_2, ..., x_N\}_t \in \mathcal{X} \subset \mathbb{R}^n$. Additionally, we define the exploration space of the robots as $\mathcal{S} \in [0,L_0]\times \ldots [0, L_{v-1}]$ where the space is bounded by $L_i$ and has a dimension of $v\leq n$. We can then construct a pointwise map $g: \mathcal{X} \to \boldsymbol{\lambda} \in \mathcal{S}$ where $\boldsymbol{\lambda}$ are the robots' trajectories. In other words, $g(x)$ maps from the state space to the exploration space.

 Next, we define control inputs to the robots as $\bold{u_t} = \{u_1, u_2, ..., u_N\}_t \in \mathcal{U} \subset \mathbb{R}^m$. The set of states of the robots over a time horizon $T$ are given as
\begin{equation}
    \bold{x}_{t+1} = f(\bold{x}_t, \bold{u}_t)
\end{equation}
where $f: \mathcal{X}\times \mathcal{U} \to \mathcal{X}$ are the dynamics of the robots. Associated with this exploration space is an information density $\Phi(\mathcal{S})$; methods for calculating information density are described later in this section. A robot's trajectory is defined to be ergodic if the robot's time-averaged statistics are proportional to the information measure $\Phi$. Formally, this means
\begin{equation} \label{eq:ergodicity}
    \lim_{T\to \infty} \frac{1}{T}\sum_{i=1}^{N}\sum_{t=0}^{T-1} h(g({x}_t)) = \sum_{i=1}^{N}\int_\mathcal{S} \Phi(\lambda)h(g(\lambda) d\lambda
\end{equation}
for all Lebesque integrable functions $h \in \mathcal{L}^1$ \cite{de2016ergodic}.

To quantify the ergodicity of a set of trajectories $\boldsymbol{\lambda}$, we consider the time-averaged statistics $C_k$ of $\boldsymbol{\lambda}$:
\begin{equation} \label{eq:time_avg_stat}
c_k(x) = \frac{1}{T} \sum_{i=1}^{N}\int_0^TF_k(g(x_i(t))dt
\end{equation}
where $F_k(\lambda) = \prod_{i=0}^{v-1} \cos(\lambda_i k_i \pi/ L_i)/h_k$ is the cosine Fourier transform for the $k^\text{th}$ mode and $h_k$ is a normalization factor \cite{7350162}.

To drive trajectories to ergodicity, we optimize over $\boldsymbol{\lambda}_{0:T-1}$ and control inputs $\bold{u}_{0:T-1}$ to minimize an error metric called the ergodic metric, defined as

\begin{align}
    &\mathcal{E}(\bold{x}, \Phi) = \sum_{k\in \mathbb{N}^v} \Lambda_k \left( c_k(\bold{x}) - \int_{\mathcal{S}} \Phi(\boldsymbol{\lambda}) F_k(\boldsymbol{\lambda})d\boldsymbol{\lambda} \right)^2
\end{align}
where $\Lambda_k=\frac{1}{1+\norm{k}^{\frac{v+1}{2}}}$ are weights assigned to the Fourier coefficients $F_k$. Finally, to find the set of states and controls which minimize the ergodic metric and while adhering to the system's control limitations, we minimize over the augmented cost function $\mathcal{J}$:
\begin{align} \label{eq:erg_opt}
    & \min_{\mathbf{x}, \mathbf{u}}\mathcal{J}(\bold{x}, t)= \mathcal{E}(\bold{x}, \Phi) + \sum_{t=0}^{T-1} \bold{u}_t^\top R \bold{u}_t \\ 
    &\text{subject to } \bold{x}_{t+1} = f(\bold{x}_t, \bold{u}_t), \bold{x}_t \in \mathcal{X}, \bold{u}_t \in \mathcal{U} \nonumber \\ 
    & \bold{x}_0 = \bar{\bold{x}}_0, g(\bold{x}_t) \in \mathcal{S}  \nonumber
\end{align}
where $\bar{\bold{x}}_0$ is the initial set of states of the robots and $R$ is a weight controlling the relative priority of minimizing the control input to the system. The planned trajectories $\boldsymbol{\lambda}$ collectively minimize the ergodic metric, meaning the union of all trajectories of the multi-agent team tends towards ergodicity. Additionally, all planned trajectories $\boldsymbol{\lambda}$ are feasible as the dynamics of each robot acts as a constraint in the optimization problem. Eq.~\eqref{eq:erg_opt} is solved using an interior point constraint solver~\cite{boyd2004convex,lukvsan2004interior}.

\subsection{Expected Information under Constrained Sensor Performance}
To begin, we define the uncertainty distribution in target states across the exploration space $S$ to be $\mathcal{V}$. We assume $\mathcal{V}$ is known at all times in the exploration process. Given an environmental process model $E(s)$ for $s \in S$, we can determine a sensor performance model $M$ where $M\big(E(s)\big) \in [0, 1]$ This sensor performance model evaluates $E$ at a given location of the exploration space, then returns a coefficient encoding the predicted sensor effectiveness at that location given the environmental process. A coefficient of 0 represents that the environmental process completely inhibits the sensor from capturing information. We can construct a preliminary EID, $\Phi_{ 1}$, by weighting $\mathcal{V}$ by the sensor effectiveness coefficients $M$:
\begin{equation}
    \Phi_{1} = \mathcal{V} \odot M
\end{equation}
where the $\odot$ operation represents pointwise matrix multiplication. 

A more advanced EID, $\Phi_{2}$, can be determined through the use of Shannon entropy. In this application, Shannon entropy is the average amount of information gain (via uncertainty reduction) associated with a given measurement. Our motivation for using Shannon entropy is to obtain an information distribution which not only considers sensor effectiveness (like $\Phi_{1}$), but also considers and compensates for sensor noise. First, we construct a probability density function (PDF)
\begin{equation}
    P(d_s, m_s) = \exp 
    \left[ \frac{-\mathcal{V}(s)\Big(\frac{d_s}{\mathcal{V}(s)}-m_s\Big)^2}{a}\right]
\end{equation}
where $d_s$ is a sensor measurement at a point $s$, $m_s = M(s)$, and $a$ is an experimentally-determined scalar that controls spread. Essentially, this PDF returns the likelihood of achieving the maximum possible uncertainty reduction, $\mathcal{V}(s)$, when measuring at a point $s$. The likelihood of achieving maximal uncertainty reduction is affected both by the sensor performance coefficient $m_s$, which dictates to what degree the environmental phenomena will obstruct sensor measurements; and the ratio $\frac{d_s}{V(s)}$, which encodes how much the measured value $d_s$ varies from the true uncertainty value $V(s)$ due to noise. For example, for a measurement $d_s$ close to $\mathcal{V}(s)$ and a sensor coefficient $m_s=1$, the PDF will return a likelihood that is close to 1; physically, this means the measurement is unobstructed and minimally noisy, and thereby is likely to yield near maximal reductions in uncertainty. 

The set of potential measurements at a point $s$ is written as $D_s$, where each value $d_s \in D_s$ is sampled from the normal distribution $\mathcal{N}(\mathcal{V}(s), \sigma ^2)$. The Gaussian noise model is applied to simulate the effect of sensor noise on the informativeness of a measurement. We note that sensor noise is distinct from the sensor performance model; sensor noise is an internal process, while the sensor performance is externally imposed from an environmental phenomena. To account for sensor noise before calculating Shannon entropy, we apply a Gaussian filter to $D_s$. Then, we can calculate the Shannon entropy at $s$ to be
\begin{equation}
    H(D_s) = -\sum_{d_s\in D_s}P(d_s, m_s)\log_2 P(d_s, m_s)
\end{equation}
Finally, we can calculate the second EID $\Phi_{2}$ to be
\begin{equation}
    \Phi_{2} = H(D_S)
\end{equation}

\section{ERGODIC EXPLORATION OF A WILDFIRE}
In this section, we describe the specific details of using multi-agent ETO to explore a wildfire. We first discuss how wildfire smoke data is simulated, and how we then explain how this data is used in the ETO algorithm.
\subsection{Smoke Modeling}
Smoke simulations were obtained using the Jet framework from \cite{kim2017fluid}. The motion of any constant-density fluid is described by the incompressible Navier-Stokes equation:
\begin{align}\label{eq:navier_stokes}
    & \rho\frac{D\bold{V}}{D\bold{t}} = \rho\bold{g}-\boldsymbol{\nabla}p + \mu\boldsymbol{\nabla}^2\bold{V} \\
    & \nabla\cdot\bold{V}=0 
    \nonumber
\end{align}
where $\bold{V}$ is velocity, $\rho$ is density, $p$ is pressure, and $\mu$ is viscosity. As described in \cite{kim2017fluid}, fluid is modeled as a multidimensional grid, the cells of which store information about the fluid's physical properties. To simulate the motion of a fluid, the gravity, pressure, and viscosity forces of Eq. \eqref{eq:navier_stokes} are calculated and used to update the fluid's velocity field $\bold{u}$. To transfer data along grid points as the simulation evolves (a process called advection), grid points are traced backwards by linearly interpolating on nearby grid values, yielding
\begin{equation}\label{eq:advection}
f(\bold{s})^{n+1} - \Tilde{f}(\bold{s}-\Delta t\bold{u})^n
\end{equation}
for a quantity $f$ at time step $n$ and position $\bold{s}$.

To simulate smoke, one additional force must be considered. Due to differences in the temperature of the smoke and the surrounding air, smoke undergoes a buoyancy force, which is approximated as
\begin{equation}
\bold{f}_{b} = -\alpha\rho\bold{y} + \beta(T-T_{amb})\bold{y}
\end{equation}
where $\alpha, \beta$ are scaling factors, $T$ is temperature, and $\bold{y}$ is vertical position. Adding this buoyant force into Eq. \eqref{eq:navier_stokes}, the velocity field $\bold{u}$ can be solved. Then, the density field $\boldsymbol{\rho}$ can be determined by applying Eq. \eqref{eq:advection} with $f(\bold{s})=\rho(\bold{s})$.

From $\boldsymbol{\rho}$, we can construct a simple model for the drones' sensor visibility. For a sensor observing the point $s$ at time $t$, the visibility coefficient of the sensor is
\begin{equation}\label{eq:visibility}
m(s, t) = 
    \begin{cases} 
      1-\frac{\boldsymbol{\rho}(s,t)}{c} & \boldsymbol{\rho}(s,t) \leq \mathrm{c} \\
      0 & \boldsymbol{\rho}(s,t) > \mathrm{c}
   \end{cases}
\end{equation}
where $\mathrm{c}$ is the cutoff density. Eq.~\eqref{eq:visibility} encodes a sensor performance model where visibility decreases linearly with smoke density until a cutoff density is reached; at that cutoff, the smoke is considered too dense for any information to be obtained, and so the visibility coefficient becomes zero.

\begin{algorithm}
 \caption{\strut Multi-Agent Visibility-Aware Ergodic Exploration}
 \begin{algorithmic}[1]
  \State \textbf{init:} $\mathcal{V}, E, N, \bold{x}, t_f, t_u$
  \State $\Phi \gets$ expectedInformation($\mathcal{V}, E$)
\State $\bold{x}, \bold{u}$ = ergodicPlanner($\bold{x}$, $\Phi$)
\While{$t<t_f$}
    \For{ $\bold{x}_{i},\bold{u}_{i} \in \bold{x}, \bold{u}$}
        \State step trajectory
        \For{$s \in \lambda_i$}
            \State $m_s \gets$ getVisibilityCoeffs()
            \State $\mathcal{V}(s) \gets (1-m_s)\mathcal{V}(s)$
        \EndFor
        \If{$t\mod{t_u}$}
            \State $E \gets$ getSmokeDen($t$)
            \State $\Phi \gets$ expectedInformation($\mathcal{V}, E$)
            \State $\bold{x}, \bold{u}$ = ergodicPlanner($\bold{x}$, $\Phi$)
        \EndIf
        \State $t \gets t + 1$
    \EndFor
\EndWhile
  \end{algorithmic} \label{alg:1}
\end{algorithm}

\begin{figure}[b!]
    \centering
    \includegraphics[width=\linewidth]{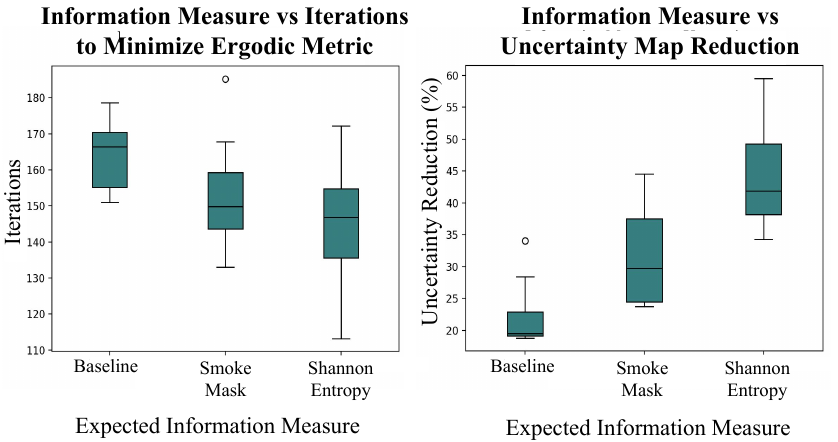}
    \caption{\textbf{Stationary Targets: Expected Information Measure Performance Comparison.} The average iterations required to minimize the ergodic metric (left) and the percent reduction in uncertainty (right) for multi-agent ETO are compared across three different approaches for calculating expected information.}
    \label{fig:fig_3}
\end{figure}

\subsection{Ergodic Exploration under Smoke-Based Visibility Occlusion}
To test our ergodic exploration method, we performed simulations of drone exploration of a wildfire on two different types of uncertainty maps. The first set of maps encoded uncertainty in the position of static targets, and the second set of maps encoded uncertainty in the position of moving targets. Both types of maps were initialized as randomly-placed Gaussian peaks of varying size and quantity. When observations were taken near the position of a target, the uncertainty peak associated with that target was reduced proportional to the visibility coefficients from Eq. \eqref{eq:visibility}. When a target moved, the uncertainty peak associated with that target was reset to its initial value as movement corresponds with a loss in confidence about the position of that target.

Smoke density data was generated offline using the Jet framework for a point source fire at random initial positions. The number of drones ranged from 2 to 10, with larger drone teams used on larger information maps. For each unique search scenario (in terms of static vs moving targets and number of drones), 10 different uncertainty maps were generated and one exploration trial was performed per map. In the simulation, measurements were collected by the drones at each step along their trajectories, and the uncertainty map was updated according to the visibility coefficient at that step.  The trajectories were then re-planned according to the current value of the EID after a specified amount of time, $t_u$, until the program reached its final time, $t_f$. The full process of our ergodic exploration method is described in Algorithm 1 and illustrated in Fig.~\ref{fig:fig_2}.

\section{RESULTS}

\begin{figure*}[t!]
    \centering
    \includegraphics[width=\linewidth]{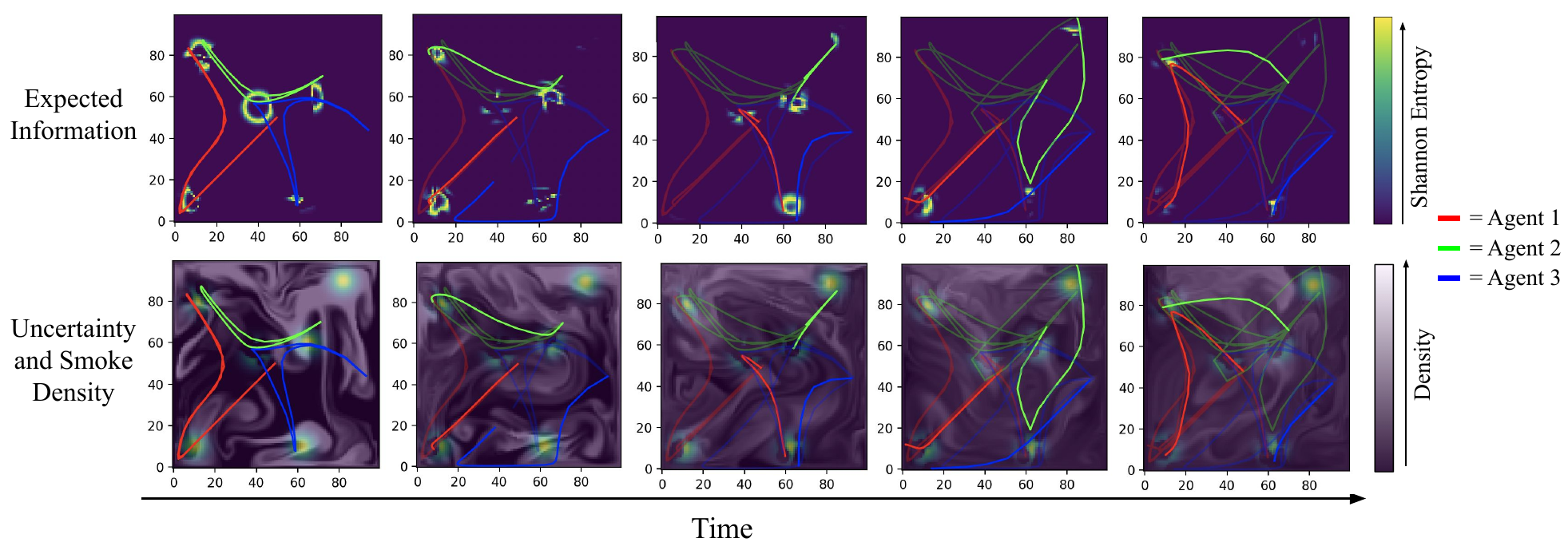}
    \captionsetup{width=\linewidth}
    \caption{\textbf{ETO Trajectory Evolution using Shannon Entropy.} For a sample uncertainty map, the EID is calculated from Shannon entropy at each time step. Ergodic trajectories are then planned at each time step over the current EID. It can be seen that at uncertainty peaks with high instantaneous smoke density, the expected information yield is low, and so the drones spend little-to-no time near those peaks. Later, as the smoke moves, some previously obstructed peaks become unobstructed and drones then visit those peaks. Over time, the total uncertainty decreases due to measurement by the drones.}
    \label{fig:fig_4}
\end{figure*}

\subsection{Stationary Targets}
To begin, we tested the performance of the multi-agent ETO algorithm on a series of randomly-generated uncertainty maps for static targets. We also compared the ETO algorithm's performance between the two EIDs, $\Phi_{1}$ and $\Phi_{2}$, described in Section III(B). These two approaches to calculating expected information are referred to as the 'smoke mask approach' and the 'Shannon entropy approach', respectively. Additionally, we evaluated the ETO algorithm on a baseline information measure, for which the uncertainty map $\mathcal{V}$ is used directly as the EID $\Phi$. In other words, the 'baseline approach' does not consider the smoke model in planning trajectories, only the uncertainty distribution. The relative performance of these three methods is shown in Fig.~\ref{fig:fig_3}. The trajectories generated by the Shannon entropy approach over time are shown in Fig. ~\ref{fig:fig_4}.

We find that, on average, the Shannon entropy approach minimizes the ergodic metric in 11.7\% and 5.42\% fewer iterations than the baseline and smoke mask approaches, respectively. This indicates the Shannon entropy approach is able to achieve ergodic coverage more rapidly than the two alternative approaches. We also find that, on average, the Shannon entropy approach reduces uncertainty by 93.8\% and 38.1\% more than the baseline and smoke mask approaches, respectively. So, the Shannon entropy approach also yields the best performance in gathering information on target states.


\begin{figure}[b!]
    \centering
    \includegraphics[width=\linewidth]{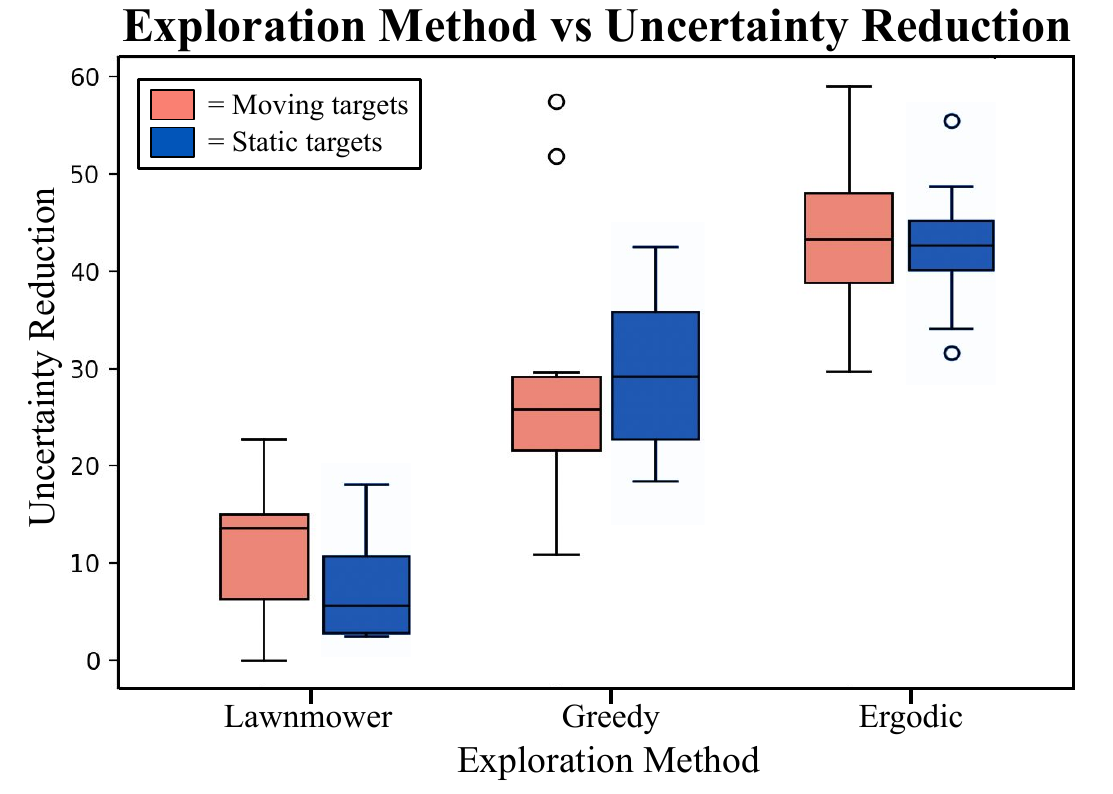}
    \caption{\textbf{Exploration Method Performance Comparison.} The percent reduction in uncertainty of our ergodic search method is compared to that of lawnmower and greedy search methods over randomly-generated uncertainty maps for both static and moving targets.}
    \label{fig:fig_5}
\end{figure}

Next, we compare the performance of our ergodic exploration method to two baseline search methods. The first is 'lawnmower search', which prioritizes coverage of the exploration map. The second is 'greedy search', which prioritizes exploitation of areas of high uncertainty. For fair comparison, the three methods are limited to the same time horizon, trajectory re-planning frequency, and maximum trajectory step size. The performance of these three methods is shown in Fig.~\ref{fig:fig_5}.

We find that our ergodic exploration method, on average, reduces 443\% more uncertainty than the lawnmower method and 40.7\% more uncertainty than the greedy method. We can also observe that the lawnmower and ergodic methods have a smaller spread than the greedy method, indicating that the greedy method has a less consistent performance at reducing uncertainty. Given that our ergodic exploration approach substantially increases uncertainty reduction and demonstrates a lower variance than the next-best method at uncertainty reduction (greedy search), while also maintaining a compute time low enough for real-time search, our approach is clearly advantageous for the problem of wildfire exploration.

\subsection{Moving Targets}
Next, we tested the performance of the multi-agent ETO algorithm on random uncertainty maps for moving targets. Further comparison of the ETO algorithm on different information measures showed that, on average, the Shannon entropy approach minimized the ergodic metric in 21.7\% and 16.9\% fewer iterations and reduced 235\% and 97.1\% more uncertainty than the baseline and smoke mask approaches, respectively. So, given that for dynamic targets, the Shannon entropy approaches continues to achieve ergodic coverage more efficiently than alternative approaches while also gathering the most information, we proceed with using the Shannon entropy approach to calculate the EID for our method. The performance of our ergodic exploration method is then compared to lawnmower and greedy exploration methods in Fig. \ref{fig:fig_5}. 


We find that our ergodic exploration method, on average, reduces 275\% more uncertainty than the lawnmower method and 51.9\% more uncertainty than the greedy method. Once again, we find that the lawnmower and ergodic methods have less spread than the greedy method. So, as with static targets, we find that our ergodic search approach outperforms the two baseline methods in reducing uncertainty while exhibiting more consistency in performance than the next-most successful exploration method of greedy search, indicating that our exploration method is advantageous for wildfire exploration for moving targets.

\section{CONCLUSIONS}
In this paper, we introduced an algorithm for multi-agent ETO that accounts for time-varying sensor visibility occlusions. We then tested this method on an evolving map of uncertainty of the state of static and dynamic targets. We demonstrated that ETO outcomes, as measured by speed and uncertainty reduction, are substantially improved by calculating expected information from the sensor visibility model. The main algorithm of this work could be applied to multi-agent exploration of any extreme environment which is dynamic in information distribution and/or sensor capabilities. In future work, we plan to test this algorithm on entropy maps derived from real-world wildfire datasets. Additionally, we are interested in extending this exploration method to heterogeneous multi-agents teams. 

\bibliographystyle{IEEEtran}
\bibliography{reference}
\end{document}